\documentclass[10pt,twocolumn,letterpaper]{article}

\usepackage[pagenumbers]{iccv} 

\definecolor{iccvblue}{rgb}{0.21,0.49,0.74}
\usepackage[pagebackref,breaklinks,colorlinks,allcolors=iccvblue]{hyperref}

\title{I2V3D: Controllable image-to-video generation with 3D guidance}

\author{
Zhiyuan Zhang$^{1}$, Dongdong Chen$^{2}$, Jing Liao$^1$$^\ddagger$  \\
$^1$City University of Hong Kong \qquad
$^2$Microsoft GenAI \qquad \\
 \vspace{-1.0em} \\ 
 Project page: \url{https://bestzzhang.github.io/I2V3D}
}

\usepackage[misc]{ifsym}
\newcommand\blfootnote[1]{
    \begingroup
    \renewcommand\thefootnote{}\footnote{#1}
    \addtocounter{footnote}{-1}
    \endgroup
}
\usepackage[hang]{footmisc}

\begin{document}
\twocolumn[{%
\maketitle
    \begin{center}
        \centering
        \captionsetup{type=figure}
        \includegraphics[width=1.0\textwidth]{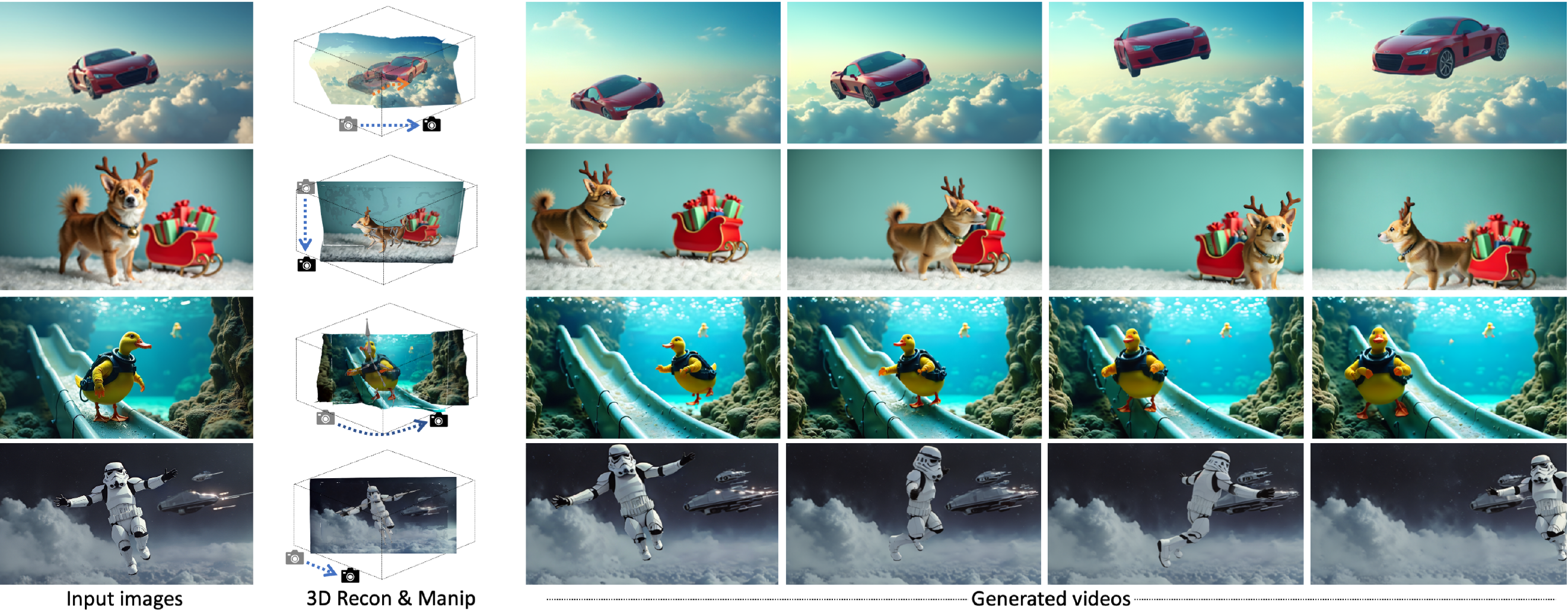}
        \captionof{figure}{
        Starting with a single image, our method reconstructs the complete scene geometry and uses the CG pipeline to enable precise control of character animation (e.g., keyframe animation or skeleton control) and camera movement (e.g., the camera rotation in the 2nd and 3rd rows, and the camera panning and zooming in the 1st and 4th rows). We then apply geometric guidance, based on the coarse rendering results, to generate high-quality, controllable videos.
        }
        \label{fig:teaser}
        \vspace{6pt}
    \end{center}%
}]{
    \blfootnote{
        $^\dagger$ Corresponding Author 
        }
}

\begin{abstract}

We present I2V3D, a novel framework for animating static images into dynamic videos with precise 3D control, leveraging the strengths of both 3D geometry guidance and advanced generative models. Our approach combines the precision of a computer graphics pipeline, enabling accurate control over elements such as camera movement, object rotation, and character animation, with the visual fidelity of generative AI to produce high-quality videos from coarsely rendered inputs. To support animations with any initial start point and extended sequences, we adopt a two-stage generation process guided by 3D geometry: 1) 3D-Guided Keyframe Generation, where a customized image diffusion model refines rendered keyframes to ensure consistency and quality, and 2) 3D-Guided Video Interpolation, a training-free approach that generates smooth, high-quality video frames between keyframes using bidirectional guidance. Experimental results highlight the effectiveness of our framework in producing controllable, high-quality animations from single input images by harmonizing 3D geometry with generative models. The code for our framework will be publicly released.
\end{abstract}    
\section{Introduction}
\label{sec:intro}

Recent advancements in image-to-video generation models~\cite{videoworldsimulators2024, esser2024scaling, xing2025dynamicrafter, lin2024open} have significantly improved the ability to produce high-quality videos with dynamic motions. However, these methods still face substantial challenges in providing precise controllability during the generation process. For example, Stable Video Diffusion (SVD)~\cite{blattmann2023stable} constrains the first frame of the generated video to be the given input image, offering no user control over subsequent frames. While subsequent text-based image-to-video methods~\cite{polyak2024movie, kong2024hunyuanvideo} allow rough motion control using textual prompts, they fall short of achieving precise control over object and camera movements as well. To address these limitations, bounding box-based~\cite{wang2024boximator, wu2024motionbooth} and trajectory-based~\cite{wu2025draganything} approaches enable users to control object movements via 2D signals such as bounding boxes or dragging gestures. However, these methods struggle with motion sequences requiring an understanding of 3D geometry and perspectives. Skeleton-based techniques~\cite{hu2024animate, chang2023magicpose} provide an alternative by using skeletal structures as control signals which, while effective for human characters, are inherently limited to human-like subjects. These constraints underscore the inability of current image-to-video generation methods to achieve fine-grained 3D control, such as rotating an object by 180 degrees, as shown by the examples of the dog and stormtrooper in Fig.~\ref{fig:teaser}.

In contrast to existing generative models, traditional Computer Graphics (CG) pipelines offer a mature and versatile set of tools—such as rigging, keyframe animation, and inverse kinematics—that enable highly flexible and precise control over object motion and camera movement in 3D space. Tools like Blender, Maya, and Houdini provide robust frameworks for animation, rigging, and simulation, making them a preferred choice for achieving detailed control. However, generating photorealistic videos with these pipelines depends on the availability of high-quality 3D models for animation and rendering. Unfortunately, existing automatic 3D model reconstruction methods from a single image~\cite{xu2024instantmesh, wang2024dust3r} remain too coarse to meet the required standards for photorealism.

In this paper, we propose a novel controllable image-to-video generation framework that combines the precise 3D control of traditional CG techniques with the photorealistic generation capabilities of advanced generative models. Given an input image, our approach consists of two main components: 1) 3D reconstruction to produce coarse 3D geometry guidance; and 2) 3D guided video generation. For the first component, our method reconstructs the entire scene in 3D from the given image. Foreground objects are extracted and converted into 3D meshes, enabling both simple and complex animations. For the background, we first inpaint occluded regions, then expand it using multi-view generation, and finally reconstruct it as a 3D mesh as well.  By reconstructing the entire scene in 3D, our method facilitates camera movement and meaningful interactions between foreground objects and their surroundings. The resulting 3D animations are then rendered into coarse outputs, including RGB frames and depth maps, which serve as 3D guidance for generating high-quality videos. 


Building upon the 3D guidance, the video generation pipeline employs a two-stage process: 1) 3D-guided keyframe generation and 2) 3D-guided video interpolation. In the first stage, we customize an image diffusion model to learn the appearance of the input image and utilize coarse renderings to guide the generation of keyframes. To enhance the generalization ability of the customized diffusion model, we augment the training process with novel-view images of the foreground object, ensuring its appearance is captured from various angles. Additionally, the temporal consistency between keyframes is further improved through extended attention mechanisms. In the second stage, we introduce a training-free, bidirectional 3D-guided interpolation method that generates high-quality, visually consistent videos between adjacent keyframes. This two-stage framework offers two key advantages. First, it allows users to define the starting point of the animation, without restricting them to the input image as the initial frame. Second, it alleviates the temporal error accumulation issue, enabling the generation of videos that extend beyond the typical window length of video diffusion models.

Extensive experiments demonstrate the superiority of our method in generating high-quality, temporally consistent videos with fine-grained controllability over camera and object motion. Our method enables users, especially professional users, to flexibly and precisely control the motions in generated videos by leveraging CG animation tools. However, compared with the traditional CG pipeline, our method has significantly lowered the professional threshold and reduced the time cost.  It automates modelling and rendering by integrating AI-based 3D reconstruction and video generation with CG animation. Thus, users only need to specify the motion to customize the animation from a single image. Our key contributions are summarized as below:

\begin{itemize}
\item We propose a novel controllable image-to-video framework that combines the strengths of both worlds, i.e.,  the fine-grained controllability of 3D graphics pipelines and the photorealistic generation capabilities of generative models.
\item We introduce a two-stage video generation pipeline with 3D guidance, comprising 3D-guided keyframe generation and video interpolation. This design allows for flexibility in specifying arbitrary starting points and generating extended video sequences. 
\item We incorporate multiple technical innovations to enhance quality, including multi-view augmentation for improved generalization, extended attention for better temporal consistency, and a bidirectional, training-free interpolation method for video synthesis.
\end{itemize}
\section{Related Works} 
\subsection{Single-View Reconstruction}
3D reconstruction is a key challenge in computer vision and graphics. Recent advances in image-to-3D generation~\cite{melas2023realfusion, hong2023lrm, tochilkin2024triposr} have improved results, especially for objects with simple backgrounds. To further enhance reconstruction quality, recent studies~\cite{li2023instant3d, tang2025lgm} have explored leveraging sparse-view inputs. For example, InstantMesh~\cite{xu2024instantmesh}, used in this paper, is a feed-forward sparse-view reconstruction model that utilizes novel views generated by Zero123++\cite{shi2023zero123++} to create 3D meshes. However, reconstructing entire scenes from a single image remains underdeveloped due to poor inpainting and warping quality\cite{chung2023luciddreamer, yu2024wonderworld, fridman2024scenescape, zhang2024text2nerf} or limited resolution~\cite{sargent2023zeronvs, szymanowicz2024flash3d}. Recent advancements~\cite{wang2023motionctrl, yu2024viewcrafter} have demonstrated improvements by employing video diffusion models for multi-view generation. In this paper, we adopt ViewCrafter~\cite{yu2024viewcrafter} for novel view generation and use a stereo model~\cite{wang2024dust3r} to extract background meshes. Despite these advancements, reconstructed 3D meshes still remain coarse and lack fine details, motivating us to leverage generative models to produce photorealistic videos.

\subsection{Image Editing with 3D Manipulation}


Manipulating 3D objects in 2D images has evolved from early methods like Cuboid Proxies~\cite{zheng2012interactive} and Stock 3D Models~\cite{kholgade20143d}, which were limited to simple shapes or predefined assets. Recent approaches fall into two categories. One leverages image diffusion models to infer object poses after manipulation without explicit 3D reconstruction. For example, 3Dit~\cite{michel2024object} fine-tunes Zero-1-to-3~\cite{liu2023zero} for 3D-aware editing but struggles with diverse objects, while Neural Assets~\cite{wu2024neural} enables multi-object pose control but is limited to rigid transformations. The other category uses 3D proxies for editing. Diffusion Handles~\cite{pandey2024diffusion} employs foreground depth but has constrained rotation, while 3DitScene~\cite{zhang20243ditscene} uses 3D Gaussian Splatting to refine 3D representations but still cannot handle non-rigid changes.

A recent work, ISculpting~\cite{yenphraphai2024image},  adopts an approach similar to ours by lifting foreground objects into 3D, manipulating them, re-rendering them into 2D, and refining the results through a coarse-to-fine process. However, there are two key differences between their method and ours. First, their approach is designed specifically for image manipulation, and directly applying it to video generation results in significant flickering issues. Second, they only reconstruct the foreground and blend it with the background after manipulation. In contrast, our method reconstructs the entire scene geometry, including both foreground and background. This enables camera movement and supports meaningful interactions between foreground objects and their surroundings, providing greater flexibility and realism.

\subsection{Controllable Video Generation}
A major challenge in video generation is enabling precise control to align videos with user intentions. Recent work has introduced various control strategies. Some methods use dense spatial conditions, such as depth maps, sketches, or canny edges, to control video generation~\cite{wang2024videocomposer, chen2023control, zhang2023controlvideo, lin2024ctrl}. Human poses are also commonly used as conditions to create dance videos~\cite{hu2024animate, chang2023magicpose, ma2024follow}. However, finding a suitable video that matches the user’s imagination and provides these control signals is often difficult.

Alternative approaches allow users to input sparse conditions, such as bounding boxes ~\cite{ma2023trailblazer, huang2023fine, wang2024boximator} or trajectories ~\cite{yin2023dragnuwa, feng2024i2vcontrol, shi2024motion, li2024image}. For example, Boximator~\cite{wang2024boximator} combines hard and soft bounding box constraints to control the position, shape, or motion path of an object. DragAnything~\cite{wu2025draganything} enables users to define motion trajectories for objects or backgrounds through simple interactions, such as selecting regions and drawing paths. Some methods also explore the integration of multiple conditions. For example, Direct-A-Video~\cite{yang2024direct} supports controlling camera movements by specifying relative panning or zooming distances and defining object motions using bounding boxes. MotionCtrl~\cite{wang2023motionctrl} allows for fine-grained motion control by utilizing camera poses for camera motion and trajectories for object motion. MotionBooth~\cite{wu2024motionbooth} introduces a framework for customizing text-to-video models using only images of a specific object and employs training-free methods to control both learned subject and camera motions at the inference time. However, since the trajectories and bounding boxes used in these methods are defined in 2D space, they struggle to accurately represent 3D motions. In addition, crafting sophisticated movements with these controls is often challenging and unintuitive. A recent work, Generative Rendering~\cite{cai2024generative}, proposes 4D-guided video generation with styles specified by text prompts. However, their method is limited to human subjects with a stationary camera and does not support generating a specific object from an image. In contrast, our approach enables users to intuitively manipulate both cameras and specified objects within a 3D engine. Furthermore, our method supports fine-grained movements, such as 3D rotations, for general objects, offering greater precision and flexibility in video generation.

\section{Method}
\begin{figure*}[t]
    \centering
    \includegraphics[width=1.0\textwidth]{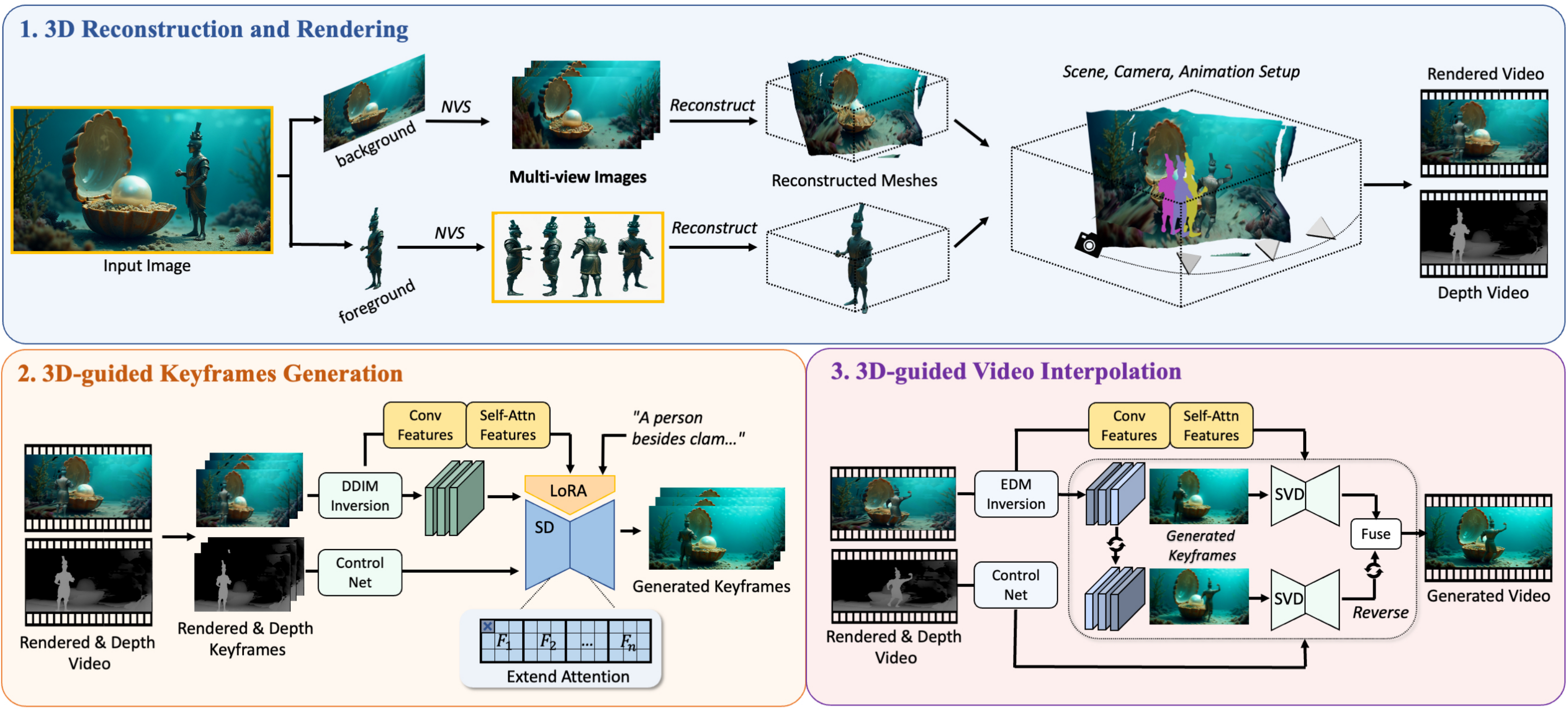}
    \caption{Our framework consists of three parts. First, we extract meshes from a single input image and use a 3D engine to create and preview a coarse animation. Next, we generate the keyframes using a 3D-guided process with an image diffusion model customized for the input image, incorporating multi-view augmentation and extended attention. Finally, we perform 3D-guided interpolation between generated keyframes to produce a high-quality, consistent video.}
    \label{fig:overall}
\end{figure*}

Given an input image, our goal is to generate an animation video starting from any desired point, guided by precise 3D control signals specified by the user. As illustrated in Fig.~\ref{fig:overall}, our proposed pipeline begins by reconstructing the entire 3D scene, creating an animation using a traditional computer graphics (CG) pipeline, and producing a high-quality video guided by 3D signals such as depth and coarsely rendered outputs. Specifically, we first extract 3D meshes for both the foreground objects and the background scene. Next, we utilize Blender, a widely adopted 3D engine, to compose the animation and render the video. Finally, we adopt a two-stage video generation process: 3D-guided keyframe generation followed by 3D-guided interpolation between keyframes. This approach provides two key benefits. First, it offers users the flexibility to define any starting frame, removing the constraint of being tied to the input image. Second, it prevents error accumulation artifacts, enabling the generation of videos that extend well beyond the temporal window typically supported by video diffusion models, ensuring both high quality and temporal consistency.

\subsection{3D Reconstruction and Rendering} \label{sec: reconstruct}

In this stage, we first lift the 2D image into a 3D scene and set up the camera and animation within a 3D engine. To create disentangled meshes, we isolate the foreground object using the object mask generated by SegAnything~\cite{kirillov2023segment} and in-paint the object’s region by using SDXL inpainting~\cite{podell2023sdxl} to obtain a complete and clean background. For the foreground mesh, we use InstantMesh~\cite{xu2024instantmesh}, a sparse-view feed-forward method that generates six novel-view images of the foreground, which are later reused in the customization stages. For the background, we begin by generating high-quality novel views and then apply Multi-View Stereo (MVS) reconstruction to obtain the 3D mesh. Specifically, we employ ViewCrafter~\cite{yu2024viewcrafter}, a state-of-the-art camera pose-conditioned video generative model, to produce novel views by setting predefined camera trajectories, such as rotating left/right or moving up/backward. A dense stereo model, Dust3r~\cite{wang2024dust3r}, is then used to reconstruct the background meshes from 1–4 of these novel-view images.

Once the meshes are generated, they are imported into a 3D engine to create animations. Users can leverage familiar 3D content creation workflows, including rigging objects (designing skeletal structures for movement), defining camera paths, and animating characters. For rigging, users can either manually design skeletal frameworks or utilize automated rigging and animation tools like Mixamo. Additionally, we provide a custom Blender script to automate the generation of camera trajectories. This script navigates through the views selected during the Multi-View Stereo (MVS) reconstruction, simplifying the process for novice users to create animations aligned with their reconstructed 3D scenes. Finally, the scene is rendered with baked textures and a depth video, providing 3D geometric signals to guide the subsequent video generation.

\subsection{3D-guided Keyframes Generation}  \label{sec: keyframe}


After acquiring the 3D geometric signals, our objective is to generate consistent keyframes that preserve the 3D geometry of the rendered frames while maintaining the visual fidelity of the input image. To retain the input image’s appearance, we begin by customizing a StableDiffusion (SD) model ~\cite{rombach2022high} using LoRA~\cite{hu2021lora}. However, fine-tuning SD with LoRA on a single image tends to overfit to the given view, limiting its ability to generalize across different views. To mitigate this issue, we customize the LoRA using multi-view images to improve generalization. To ensure that the geometry of the generated images aligns with the rendered frames, we use depth maps and coarsely rendered frames as control conditions. Furthermore, to enhance temporal consistency across the generated keyframes, we incorporate Extended Attention, introducing additional cross-frame interactions for improved coherence.

\subsubsection{LoRA Customization with Multi-View Images} \label{sec:fine-tune}
LoRA is an efficient model adaptation technique that freezes the pre-trained weight matrices \( W_0 \) and integrates additional trainable matrices \( \Delta W \), which are decomposed into two low-rank matrices \( A \) and \( B \), significantly reducing the number of trainable parameters. In our implementation, we apply LoRA for the attention layers and feedforward layers of the self-attention and cross-attention block in SD. 



During each training step, we use the input image along with a randomly selected augmented image of the foreground object from the multiple novel-view images generated by InstantMesh~\cite{xu2024instantmesh}. To facilitate learning, we create simple prompts, such as ``a \{object\} in \{environment\}'' to describe the input image, and ``a \{object\}'' to describe the foreground object in all augmented images. The training loss is defined as follows:

\vspace{-1em}
\begin{equation}
\begin{aligned}
\mathbb{E}_{z, z', \epsilon, t, c} \Big[
    &\left\| (\epsilon - \epsilon_{\theta}(z_t, c_i, t)) \right\|_2^2 \\
    &+ \lambda \left\| (\epsilon - \epsilon_{\theta}(z'_t, c_{fg}, t)) \cdot M \right\|_2^2 \Big] 
\end{aligned}
\end{equation}

\vspace{-1em}
\noindent where \( z_t \) and \( z_t' \) are the noisy latents at time step \( t \) for the input image and the augmented image, respectively. Similarly, \( c_i \) and \( c_{fg} \) are the text prompts for the input image and the foreground object, respectively. \( M \) represents the mask for the foreground object in the augmented image, \( \epsilon \) denotes the added noise, and \( \epsilon_\theta \) refers to the denoising network. \( \lambda \) is a hyperparameter that controls the weights of the augmentation loss.
\subsubsection{Geometric Guidance for Generation Control} \label{sec:geo-guidance}

Following previous research~\cite{yenphraphai2024image}, we use both depth control and rendered feature control to preserve the geometry of the rendered keyframes. First, we invert the rendered keyframes using DDIM inversion~\cite{song2020denoising} to obtain the inverted latents \( z_{inv} \). These \( z_{inv} \) latents are then used as the initial noise for keyframe generation. During each denoising step, we apply depth control through ControlNet~\cite{zhang2023adding} and inject the self-attention features \( sa \) (consists of \( q \) and \( k \)) and the convolutional features \( conv \) from the residual block, which are extracted during the inversion process. The depth control provides coarse 3D geometric information, while the rendered features (\( sa \) and \( conv \)) further complement the layout and fine-grained geometric structure of nearby planes, as shown in previous work~\cite{ku2024anyv2v}. Additionally, we mask the injection of rendered features in the invisible regions to account for imperfections in the reconstructed background mesh.

\subsubsection{Extended Attention for Consistency Enhancement} \label{sec:extendAttn}

To enhance temporal consistency between keyframes, we further leverage the extended attention mechanism proposed in prior style-aligned generation~\cite{hertz2024style} and T2V editing works~\cite{ceylan2023pix2video, geyer2023tokenflow} . This mechanism modifies the self-attention layers such that each frame not only attends to its own features but also features from other keyframes. Specifically, the Extended Attention is calculated as follows:

\vspace{-1em}
\begin{gather}
\text{ExtAttn}\left(q_i; k_{1 \ldots n}, v_{1 \ldots n}\right) = 
\mathcal{S} \left(\frac{q_i k_{1 \ldots n}^T}{\sqrt{d}}\right) \cdot v_{1 \ldots n}
\end{gather}

\noindent Here, \( q_i \) represents the query features, while \( k_{1 \ldots n} \) and \( v_{1 \ldots n} \) denote the concatenated key and value features of all frames, respectively. The function \( \mathcal{S}(\cdot) \) represents the \textit{Softmax} operation. By capturing both spatial and temporal context through this approach, the method facilitates better interaction among keyframes, encourages them to share a consistent global appearance.

\subsection{3D-guided Video Interpolation}~\label{interpolation}

\vspace{-1em}
To generate a consistent and high-quality video from keyframes, we utilize an image-to-video interpolation approach that ensures temporal coherence. Additionally, geometric guidance is applied to maintain motion alignment with the rendered video.

Inspired by ~\cite{feng2025explorative}, we adopt a dual-trajectory denoising approach to leverage bidirectional contextual information. For each denoising step, we pass the latents through two parallel denoising trajectories. One is a forward denoising trajectory, conditioned on the first frame, and the other is a time-reversed denoising trajectory, where the latents are reversed along the time dimension and conditioned on the last frame. After denoising, the outputs from the forward and time-reversed trajectories are fused using a weighted average, proportional to the distance between the first frame and the last frame.


To improve alignment with the geometry of the rendered videos, we apply similar geometric guidance during interpolation. Specifically, we use a depth ControlNet released by the community~\cite{CiaraRowles2024}, complemented with rendered feature control. First, we invert the rendered video using EDM~\cite{Karras2022edm} to obtain the initial latents. During denoising, we use rendered depth control and override convolutional and self-attention features with those obtained during inversion. These steps ensure that the interpolated video aligns better with the geometry of the rendered video while maintaining temporal consistency.

\section{Experiments}
\begin{figure*}[tbp]
  \centering
  \captionsetup{skip=-1pt}
  \includegraphics[width=1.0\linewidth]{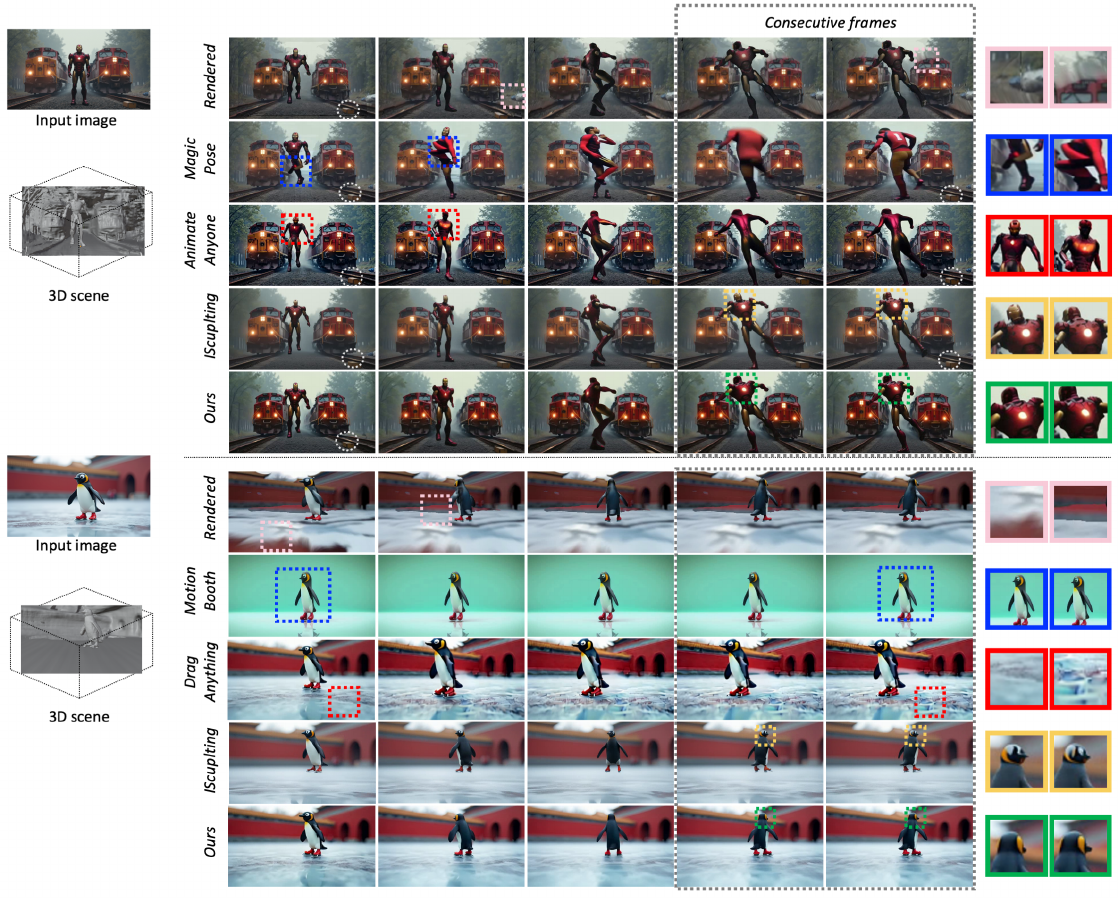}
  \caption{Qualitative comparison with baselines: (1st) human-like characters, (2nd panel) non-human objects. For human-like characters, MagicPose~\cite{chang2023magicpose} struggles with pose control (blue), and AnimateAnyone~\cite{hu2024animate} fails to preserve appearance (red). For non-human objects, MotionBooth~\cite{wu2024motionbooth} shows overfitting (blue), and DragAnything~\cite{wu2025draganything} shows error accumulation (red). ISculpting~\cite{yenphraphai2024image} exhibits frame inconsistency (yellow) for both categories. Our method outperforms them by following the geometry guidance of coarse renderings but resolves their artifacts (pink).}
  \label{fig:comparison}
  \vspace{-1.5em}
\end{figure*}

\subsection{Implementation Details}


To customize the input image, we fine-tune the SDXL model~\cite{podell2023sdxl} with LoRA rank 32 for 500 steps using a learning rate of 1e-4 and the augmentation weight $\lambda$ set to 1.0. For rendered feature control in both 3D-guided keyframe generation and video interpolation, we use self-attention features and convolutional features. For keyframe generation, which uses our customized model, we inject these features from all the upsampling blocks and set $\tau_{\text{conv}} = 0.4$ and $\tau_{\text{sa}} = 0.0$. For video interpolation, which uses the pre-trained SVD~\cite{blattmann2023stable}, we use $sa$ features from layers 4–11 and $conv$ features from layer 4, following previous training-free methods~\cite{tumanyan2023plug, ku2024anyv2v}. Here, we set $\tau_{\text{conv}}$ and $\tau_{\text{sa}}$ to 0.0. The image and video are both of the resolution 1024×576. It takes approximately 20 seconds to generate a keyframe and about 310 seconds to generate 16 video frames on a 3090Ti GPU. 
\subsection{Experiment Setup}

\subsubsection{Baselines and Dataset} We compare our method with ISculpting~\cite{yenphraphai2024image} for both human-like characters and nonhuman-like objects. Additionally, we include comparisons with two other methods in each category: 1) For human-like characters, we evaluate our method against two skeleton-driven approaches: AnimateAnyone~\cite{hu2024animate} and MagicPose~\cite{chang2023magicpose}. To use these methods, we extract skeletons from the rendered video using DwPose~\cite{yang2023effective}. These skeletons are then input to AnimateAnyone and MagicPose, which generate animations for the input images based on the skeleton data. 2) For nonhuman-like objects, we compare our method with two approaches that support both camera and object movement control. In DragAnything~\cite{wu2025draganything}, pixel trajectories are detected using Co-Track~\cite{karaev24cotracker3}. Human interaction is simulated by randomly selecting one foreground trajectory to control the object’s movement and four background trajectories to control the camera movement. In Motionbooth~\cite{wu2024motionbooth}, bounding boxes are extracted from the foreground object mask, while the camera’s movements along x and y directions are calculated by averaging the motion vectors of the background. To evaluate our methods against baselines, we created 30 animations, including 15 human-like characters and 15 nonhuman-like objects. Each animation consists of at least 61 frames (4 batches for a 16-frame diffusion model), which is the video length used in our evaluation. 


\subsubsection{Metrics} We evaluate the results from three aspects. To measure temporal consistency, we calculate the CLIP similarity~\cite{radford2021learning} between consecutively generated frames, following previous methods~\cite{esser2023structure}. For visual similarity, we use the CLIP-I score proposed in ~\cite{ruiz2023dreambooth}, which calculates the cosine similarity of the CLIP image embeddings between the generated frames and the input image. Finally, to assess the alignment between the generated and rendered videos, we employ the SSIM~\cite{wang2004image} and D-RMSE~\cite{yenphraphai2024image} metrics. D-RMSE measures the RMSE between the estimated depth of the rendered frames and the frames generated by each method, and we use DepthAnything~\cite{yang2024depth} as the depth estimator.

\begin{table}[t]
\centering
\resizebox{\columnwidth}{!}{
\begin{tabular}{l|c c c c}
\toprule
\small
\textbf{Methods} & Consistency $\uparrow$ & CLIP-I $\uparrow$ & SSIM $\uparrow$ & D-RMSE $\downarrow$ \\\hline

 & \multicolumn{4}{c}{Nonhuman-like objects} \\ 
DragAnything & 0.991 & 0.839 & 0.502 & 0.167 \\
MotionBooth & 0.974 & 0.755 & 0.519 & 0.256 \\
ISculpting & 0.986 & 0.906 & 0.570 & 0.161 \\
Ours & \textbf{0.994} & \textbf{0.906} & \textbf{0.757} & \textbf{0.102} \\ \hline \hline
& \multicolumn{4}{c}{Human-like characters} \\
AnimateAnyone & 0.985 & 0.874 & 0.407 & 0.178 \\
MagicPose & 0.952 & 0.85 & 0.425 & 0.210 \\
ISculpting & 0.982 & \textbf{0.916} & 0.507 & 0.155 \\
Ours & \textbf{0.991} & 0.901 & \textbf{0.735} & \textbf{0.117} \\ \bottomrule
\end{tabular}
}
\caption{Quantitative comparison with DragAnything~\cite{wu2025draganything}, MotionBooth~\cite{wu2024motionbooth}, ISculpting~\cite{yenphraphai2024image} on Nonhuman-like objects and AnimateAnyone~\cite{hu2024animate}, MagicPose~\cite{chang2023magicpose} and ISculpting~\cite{yenphraphai2024image} on Human-like characters.}
\label{table:quantitative}
\vspace{-1.5em}
\end{table}




\subsection{Qualitative and Quantitative Comparison}

Fig.~\ref{fig:comparison} compares our method with several baselines. For human-like videos, MagicPose~\cite{chang2023magicpose} and AnimateAnyone~\cite{hu2024animate} struggle to accurately follow reference poses or preserve appearances, as their training datasets primarily feature faces and upper bodies, while ISculpting~\cite{yenphraphai2024image} produces flickering frames. Additionally, none of these methods can handle camera movement (e.g., the box highlighted with a white circle remains static despite the camera moving). For nonhuman-like objects, MotionBooth~\cite{wu2024motionbooth} overfits, resulting in nearly static videos, while DragAnything~\cite{wu2025draganything} is limited to 2D movements and accumulates color errors. ISculpting~\cite{yenphraphai2024image} again exhibits inconsistencies in appearance. In contrast, our method successfully animates both human-like characters and nonhuman-like objects according to the reference motion, preserves the original appearance, and achieves superior temporal consistency.  

Beyond qualitative improvements, our method also excels in quantitative evaluations. As shown in Tab.~\ref{table:quantitative}, it outperforms baselines in temporal consistency (0.994 and 0.991) and structural similarity (0.102 and 0.117 for D-RMSE; 0.757 and 0.735 for SSIM). Additionally, it achieves comparable visual similarity (around 0.90) to the input image, similar to ISculpting~\cite{yenphraphai2024image}, which directly blends the input image background during denoising. These results demonstrate our method’s ability to generate temporally consistent animations that closely align with both the input image and the rendered video. Apart from the quantitative evaluations, we also conduct a user study to evaluate the effectiveness of our approach, which is included in our supplementary A.

\subsection{Ablation study}



\begin{figure}[tbp]
  \centering
  \begin{minipage}[t]{0.48\textwidth}
    \centering
    \includegraphics[width=\linewidth]{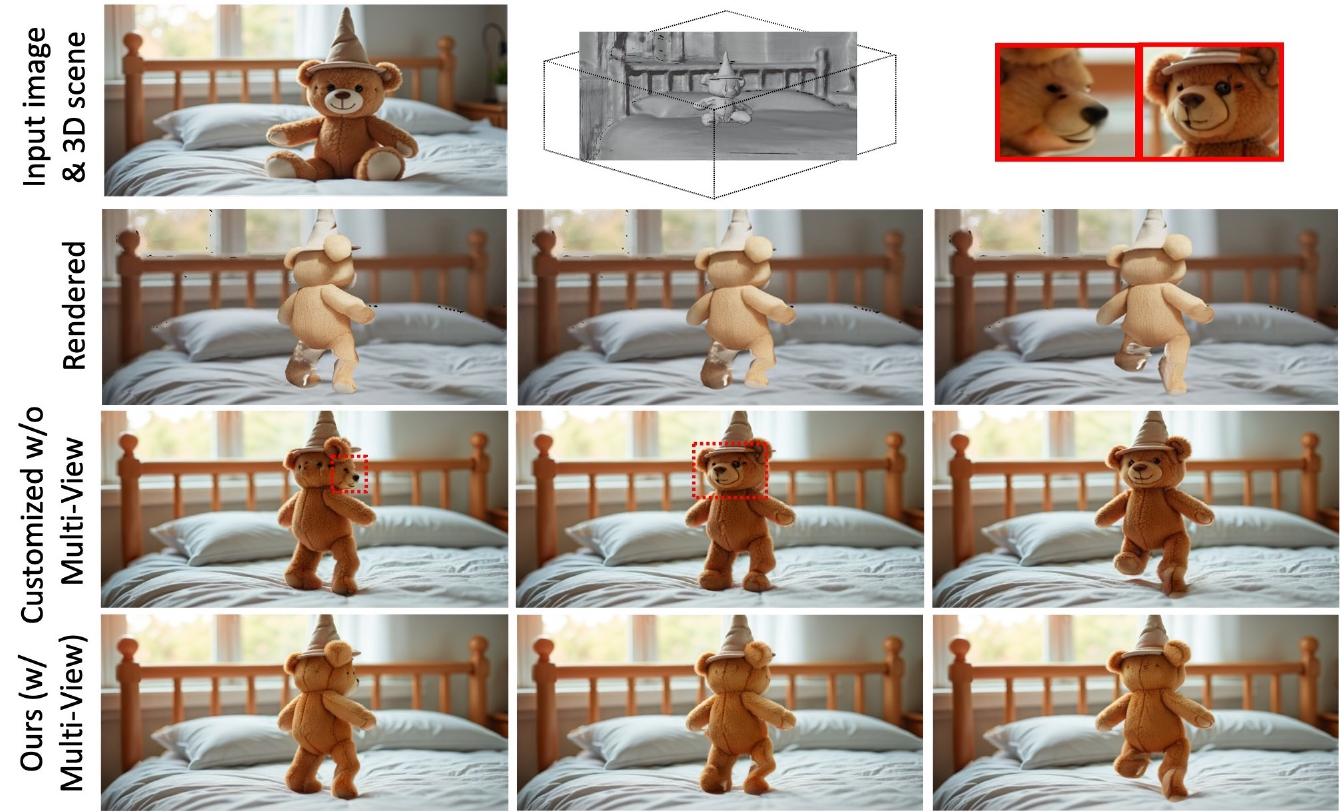}
    \captionsetup{skip=1pt} 
    \caption{Ablation on LoRA customization with multi-view image augmentation. The red boxes highlight overfitting to the frontal view.}
    \label{fig:abl_multiview}
  \end{minipage}
  
  \vspace{0.5em} 

  \begin{minipage}[t]{0.48\textwidth}
    \centering
    \includegraphics[width=\linewidth]{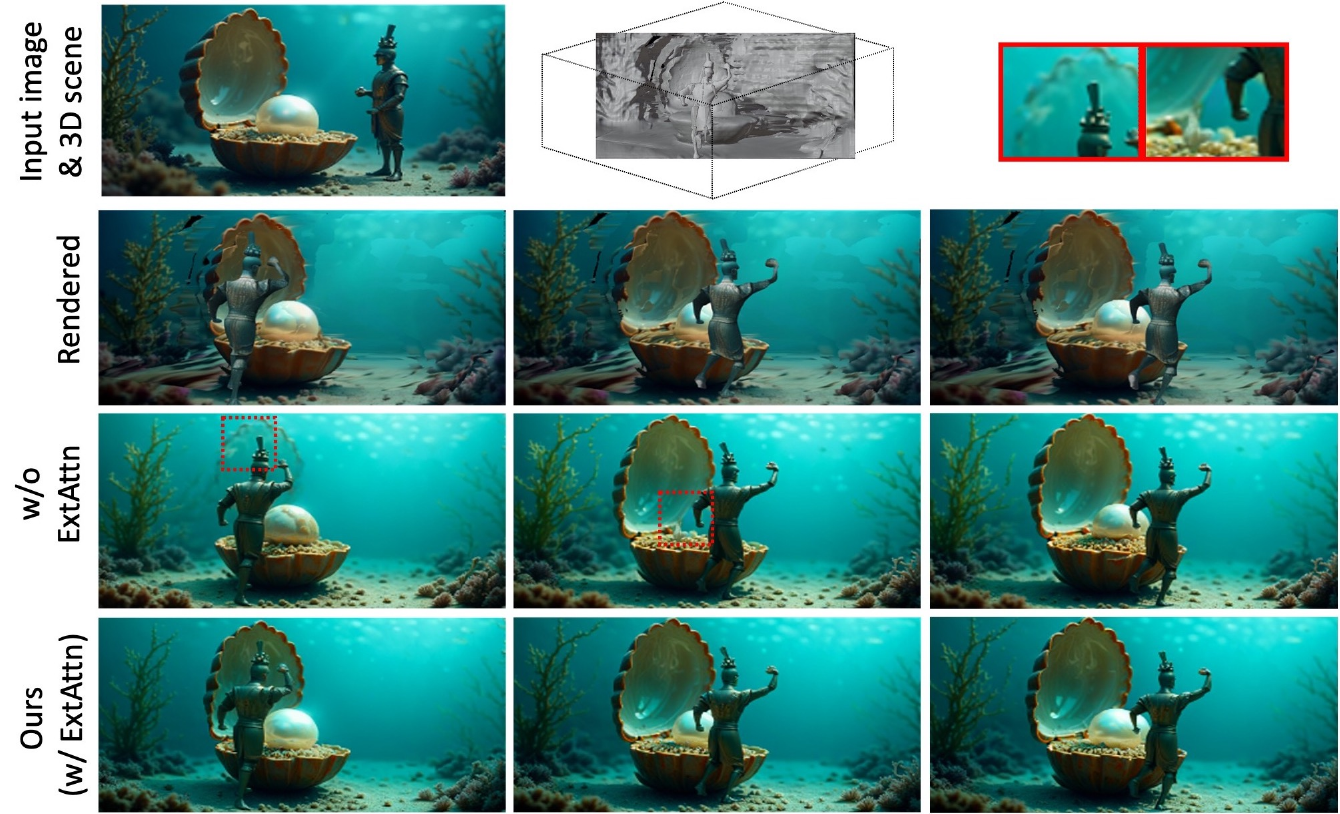}
    \captionsetup{skip=1pt} 
    \caption{Ablation on extended attention for consistency enhancement. The red boxes highlight inconsistencies between individual generated frames.}
    \label{fig:abl_extendAttn}
  \end{minipage}
\end{figure}

\begin{figure}[tbp]
  \centering
  \captionsetup{skip=3pt}
  \includegraphics[width=1.0\linewidth]{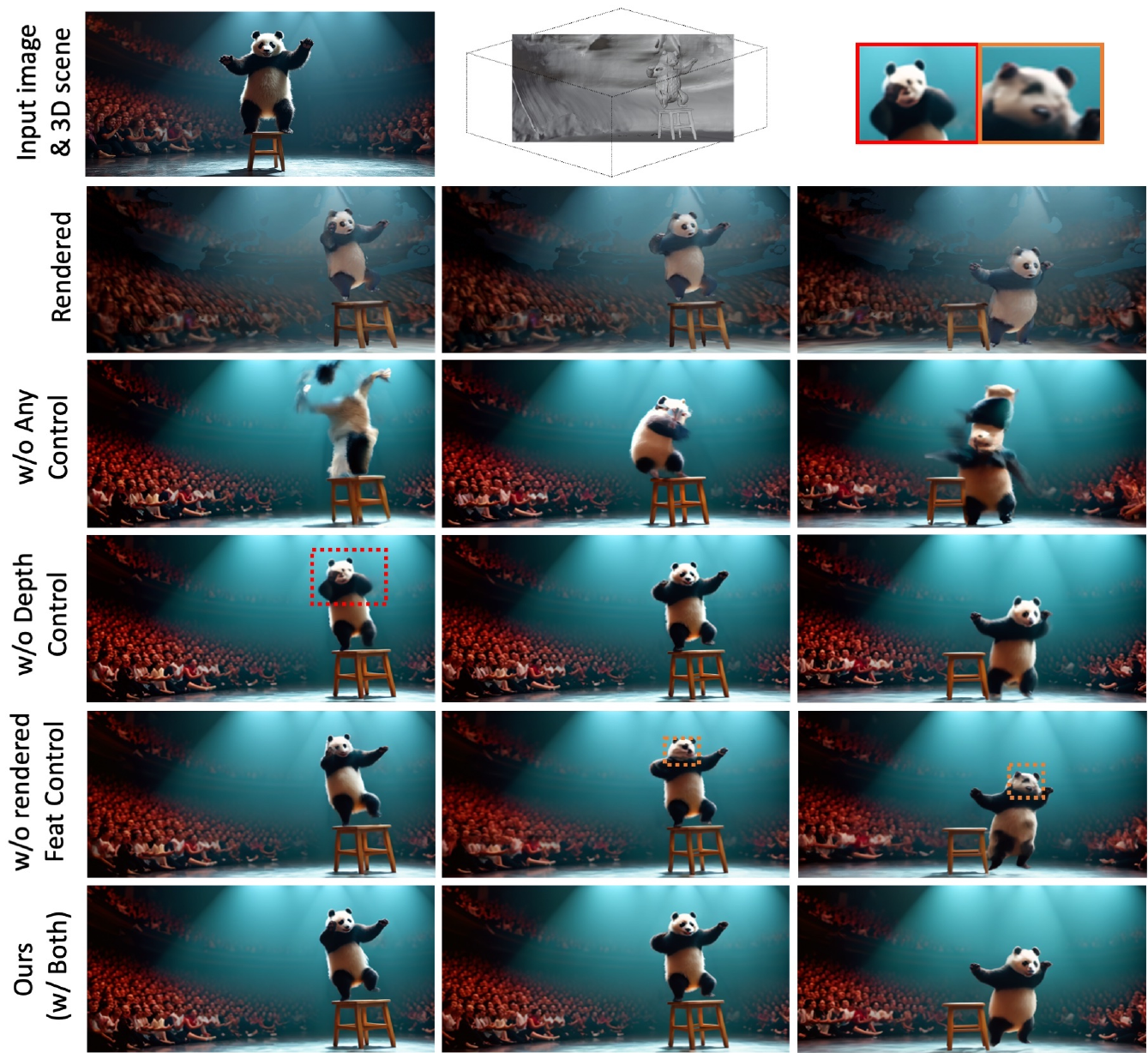}
  \caption{Ablation on 3D-guided video interpolation.}
  \label{fig:abl_interp}
  \vspace{-1.5em}
\end{figure}


\noindent\textbf{Ablation on Multi-View Customization.} Before keyframe generation, our method customizes the image diffusion model using multi-view images of the foreground object for training augmentation. As shown in the third row of Fig.~\ref{fig:abl_multiview}, customizing LoRA on a single input image overfits by generating only the frontal view when the object rotates to its back, losing the ability to generalize to diverse unseen views. In contrast, our multi-view customization effectively captures the object's appearance from different angles and generalizes much better.

\noindent\textbf{Ablation on Extended Attention.} As discussed in Sec.\ref{sec:extendAttn}, we utilize extended attention to share global appearance information between keyframes. Generating frames individually often results in inconsistencies, such as the incomplete shell and missing pearl in the third row of Fig.\ref{fig:abl_extendAttn}. In contrast, our method enhances consistency by generating the keyframes jointly with Extended Attention.

\noindent\textbf{Ablation on 3D-Guided Interpolation.} As discussed in Sec.\ref{sec:geo-guidance}, ControlNet\cite{zhang2023adding} directly captures 3D geometric information, while the rendered features enhance layout and fine-grained structure. As shown in third row of Fig.~\ref{fig:abl_interp}, interpolation without any guidance fails to generate meaningful content. When only rendered feature control is applied without depth control, the model lacks direct geometric control, as illustrated by the panda’s hand highlighted in the red box. Conversely, without rendered feature control, the objects fail to accurately capture detailed features within the same plane, such as the panda's eyes highlighted by the orange box. In contrast, our method combines both depth control and rendered feature control, achieving better alignment with the rendered video. Apart from these ablations, more ablations are included in our supplementary B.

\section{Dicussion}
\subsection{Balancing the 3D Control and Video Prior}
Designed mainly for professional CG users, our framework aims to offer fine-grained animation control while streamlining the production process by automating modeling and rendering. Although our showcased animations primarily focus on object animations, users have the flexibility to incorporate additional dynamic effects like fluid simulations and cloth animations into the 3D scene, which are compatible with our pipeline.
Alternatively, users can adjust the intensity of 3D guidance (by modifying feature injection and ControlNet scale) to leverage more motion priors of the video generation model to increase dynamism. As demonstrated in our supplementary videos, gradually reducing the 3D guidance from 1.0 to 0.5 and then to 0.2 increases the extent of dynamic motions in the background, such as water waves and flag flutters.
\begin{figure}[tbp]
  \centering
  \includegraphics[width=1.0\linewidth]{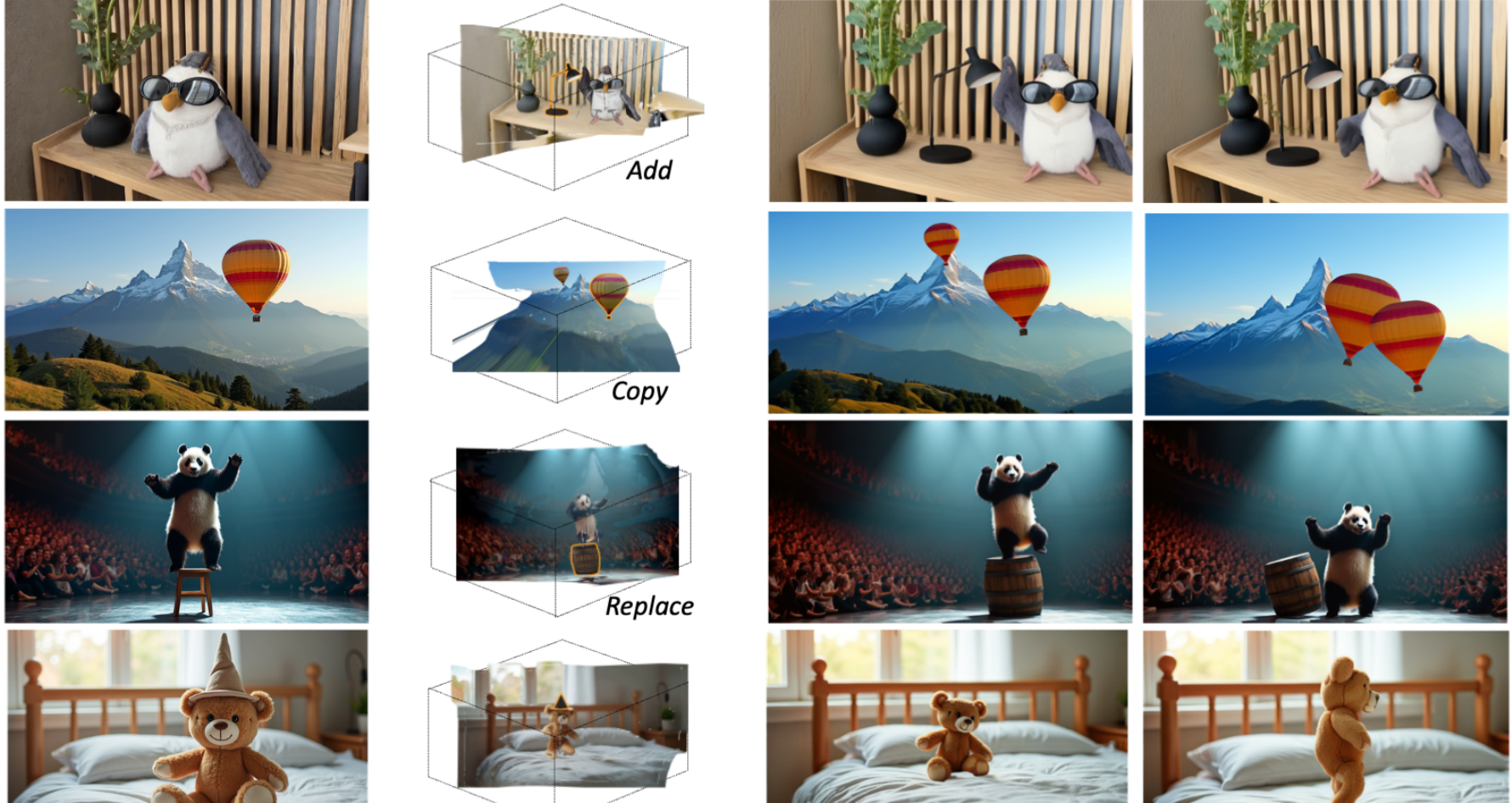}
  \caption{More applications of our method include adding, copying, replacing, and editing objects in 3D scenes to guide video generation.}
  \label{fig:more}
\end{figure}
\subsection{Video Generation in New Composed Scene}
By introducing 3D geometry control for image-to-video generation, it provides users with greater freedom to specify video content, in addition to object animation and camera movement. As demonstrated in Fig.~\ref{fig:more}, users can further add, copy, replace, and edit 3D objects to compose a new 3D scene that guides video generation.

\section{Conclusion}


In this paper, we present a novel solution to the longstanding challenge of precise and controllable video generation from a static image. By integrating 3D modeling tools and advanced generative models, our framework bridges the connection between traditional computer graphics and modern diffusion-based synthesis, enabling precise 3D control in the image-to-video generation process. Our proposed framework can achieve high-quality, temporally consistent animations while maintaining control over object and camera movements in 3D space. Experimental results validate the effectiveness of our approach, highlighting its general applicability to a wide range of scenarios, from moving and rotating objects, animating characters, to editing or composing new scenes in video generation.


{
    \small
    \bibliographystyle{ieeenat_fullname}
    \bibliography{reference}

\begin{thebibliography}{70}
\providecommand{\natexlab}[1]{#1}
\providecommand{\url}[1]{\texttt{#1}}
\expandafter\ifx\csname urlstyle\endcsname\relax
  \providecommand{\doi}[1]{doi: #1}\else
  \providecommand{\doi}{doi: \begingroup \urlstyle{rm}\Url}\fi

\bibitem[Cia(2024)]{CiaraRowles2024}
temporal-controlnet-depth-svd-v1, 2024.

\bibitem[Blattmann et~al.(2023)Blattmann, Dockhorn, Kulal, Mendelevitch, Kilian, Lorenz, Levi, English, Voleti, Letts, et~al.]{blattmann2023stable}
Andreas Blattmann, Tim Dockhorn, Sumith Kulal, Daniel Mendelevitch, Maciej Kilian, Dominik Lorenz, Yam Levi, Zion English, Vikram Voleti, Adam Letts, et~al.
\newblock Stable video diffusion: Scaling latent video diffusion models to large datasets.
\newblock \emph{arXiv preprint arXiv:2311.15127}, 2023.

\bibitem[Brooks et~al.(2024)Brooks, Peebles, Holmes, DePue, Guo, Jing, Schnurr, Taylor, Luhman, Luhman, Ng, Wang, and Ramesh]{videoworldsimulators2024}
Tim Brooks, Bill Peebles, Connor Holmes, Will DePue, Yufei Guo, Li Jing, David Schnurr, Joe Taylor, Troy Luhman, Eric Luhman, Clarence Ng, Ricky Wang, and Aditya Ramesh.
\newblock Video generation models as world simulators.
\newblock 2024.

\bibitem[Cai et~al.(2024)Cai, Ceylan, Gadelha, Huang, Wang, and Wetzstein]{cai2024generative}
Shengqu Cai, Duygu Ceylan, Matheus Gadelha, Chun-Hao~Paul Huang, Tuanfeng~Yang Wang, and Gordon Wetzstein.
\newblock Generative rendering: Controllable 4d-guided video generation with 2d diffusion models.
\newblock In \emph{Proceedings of the IEEE/CVF Conference on Computer Vision and Pattern Recognition}, pages 7611--7620, 2024.

\bibitem[Ceylan et~al.(2023)Ceylan, Huang, and Mitra]{ceylan2023pix2video}
Duygu Ceylan, Chun-Hao~P Huang, and Niloy~J Mitra.
\newblock Pix2video: Video editing using image diffusion.
\newblock In \emph{Proceedings of the IEEE/CVF International Conference on Computer Vision}, pages 23206--23217, 2023.

\bibitem[Chang et~al.(2023)Chang, Shi, Gao, Xu, Fu, Song, Yan, Zhu, Yang, and Soleymani]{chang2023magicpose}
Di Chang, Yichun Shi, Quankai Gao, Hongyi Xu, Jessica Fu, Guoxian Song, Qing Yan, Yizhe Zhu, Xiao Yang, and Mohammad Soleymani.
\newblock Magicpose: Realistic human poses and facial expressions retargeting with identity-aware diffusion.
\newblock In \emph{Forty-first International Conference on Machine Learning}, 2023.

\bibitem[Chen et~al.(2023)Chen, Ji, Wu, Wu, Xie, Li, Xia, Xiao, and Lin]{chen2023control}
Weifeng Chen, Yatai Ji, Jie Wu, Hefeng Wu, Pan Xie, Jiashi Li, Xin Xia, Xuefeng Xiao, and Liang Lin.
\newblock Control-a-video: Controllable text-to-video generation with diffusion models.
\newblock \emph{arXiv preprint arXiv:2305.13840}, 2023.

\bibitem[Chung et~al.(2023)Chung, Lee, Nam, Lee, and Lee]{chung2023luciddreamer}
Jaeyoung Chung, Suyoung Lee, Hyeongjin Nam, Jaerin Lee, and Kyoung~Mu Lee.
\newblock Luciddreamer: Domain-free generation of 3d gaussian splatting scenes.
\newblock \emph{arXiv preprint arXiv:2311.13384}, 2023.

\bibitem[Esser et~al.(2023)Esser, Chiu, Atighehchian, Granskog, and Germanidis]{esser2023structure}
Patrick Esser, Johnathan Chiu, Parmida Atighehchian, Jonathan Granskog, and Anastasis Germanidis.
\newblock Structure and content-guided video synthesis with diffusion models.
\newblock In \emph{Proceedings of the IEEE/CVF International Conference on Computer Vision}, pages 7346--7356, 2023.

\bibitem[Esser et~al.(2024)Esser, Kulal, Blattmann, Entezari, M{\"u}ller, Saini, Levi, Lorenz, Sauer, Boesel, et~al.]{esser2024scaling}
Patrick Esser, Sumith Kulal, Andreas Blattmann, Rahim Entezari, Jonas M{\"u}ller, Harry Saini, Yam Levi, Dominik Lorenz, Axel Sauer, Frederic Boesel, et~al.
\newblock Scaling rectified flow transformers for high-resolution image synthesis.
\newblock In \emph{Forty-first International Conference on Machine Learning}, 2024.

\bibitem[Feng et~al.(2025)Feng, Ding, Xia, Niklaus, Abrevaya, Black, and Zhang]{feng2025explorative}
Haiwen Feng, Zheng Ding, Zhihao Xia, Simon Niklaus, Victoria Abrevaya, Michael~J Black, and Xuaner Zhang.
\newblock Explorative inbetweening of time and space.
\newblock In \emph{European Conference on Computer Vision}, pages 378--395. Springer, 2025.

\bibitem[Feng et~al.(2024)Feng, Qi, Liu, Sun, Tu, Ma, Dai, Zhao, Zhou, and He]{feng2024i2vcontrol}
Wanquan Feng, Tianhao Qi, Jiawei Liu, Mingzhen Sun, Pengqi Tu, Tianxiang Ma, Fei Dai, Songtao Zhao, Siyu Zhou, and Qian He.
\newblock I2vcontrol: Disentangled and unified video motion synthesis control.
\newblock \emph{arXiv preprint arXiv:2411.17765}, 2024.

\bibitem[Fridman et~al.(2024)Fridman, Abecasis, Kasten, and Dekel]{fridman2024scenescape}
Rafail Fridman, Amit Abecasis, Yoni Kasten, and Tali Dekel.
\newblock Scenescape: Text-driven consistent scene generation.
\newblock \emph{Advances in Neural Information Processing Systems}, 36, 2024.

\bibitem[Geyer et~al.(2023)Geyer, Bar-Tal, Bagon, and Dekel]{geyer2023tokenflow}
Michal Geyer, Omer Bar-Tal, Shai Bagon, and Tali Dekel.
\newblock Tokenflow: Consistent diffusion features for consistent video editing.
\newblock \emph{arXiv preprint arXiv:2307.10373}, 2023.

\bibitem[Hertz et~al.(2024)Hertz, Voynov, Fruchter, and Cohen-Or]{hertz2024style}
Amir Hertz, Andrey Voynov, Shlomi Fruchter, and Daniel Cohen-Or.
\newblock Style aligned image generation via shared attention.
\newblock In \emph{Proceedings of the IEEE/CVF Conference on Computer Vision and Pattern Recognition}, pages 4775--4785, 2024.

\bibitem[Hong et~al.(2023)Hong, Zhang, Gu, Bi, Zhou, Liu, Liu, Sunkavalli, Bui, and Tan]{hong2023lrm}
Yicong Hong, Kai Zhang, Jiuxiang Gu, Sai Bi, Yang Zhou, Difan Liu, Feng Liu, Kalyan Sunkavalli, Trung Bui, and Hao Tan.
\newblock Lrm: Large reconstruction model for single image to 3d.
\newblock \emph{arXiv preprint arXiv:2311.04400}, 2023.

\bibitem[Hu et~al.(2021)Hu, Shen, Wallis, Allen-Zhu, Li, Wang, Wang, and Chen]{hu2021lora}
Edward~J Hu, Yelong Shen, Phillip Wallis, Zeyuan Allen-Zhu, Yuanzhi Li, Shean Wang, Lu Wang, and Weizhu Chen.
\newblock Lora: Low-rank adaptation of large language models.
\newblock \emph{arXiv preprint arXiv:2106.09685}, 2021.

\bibitem[Hu(2024)]{hu2024animate}
Li Hu.
\newblock Animate anyone: Consistent and controllable image-to-video synthesis for character animation.
\newblock In \emph{Proceedings of the IEEE/CVF Conference on Computer Vision and Pattern Recognition}, pages 8153--8163, 2024.

\bibitem[Huang et~al.(2023)Huang, Su, Sun, Jiang, Jia, Zhu, and Yang]{huang2023fine}
Hsin-Ping Huang, Yu-Chuan Su, Deqing Sun, Lu Jiang, Xuhui Jia, Yukun Zhu, and Ming-Hsuan Yang.
\newblock Fine-grained controllable video generation via object appearance and context.
\newblock \emph{arXiv preprint arXiv:2312.02919}, 2023.

\bibitem[Karaev et~al.(2024)Karaev, Makarov, Wang, Neverova, Vedaldi, and Rupprecht]{karaev24cotracker3}
Nikita Karaev, Iurii Makarov, Jianyuan Wang, Natalia Neverova, Andrea Vedaldi, and Christian Rupprecht.
\newblock Cotracker3: Simpler and better point tracking by pseudo-labelling real videos.
\newblock In \emph{Proc. {arXiv:2410.11831}}, 2024.

\bibitem[Karras et~al.(2022)Karras, Aittala, Aila, and Laine]{Karras2022edm}
Tero Karras, Miika Aittala, Timo Aila, and Samuli Laine.
\newblock Elucidating the design space of diffusion-based generative models.
\newblock In \emph{Proc. NeurIPS}, 2022.

\bibitem[Kholgade et~al.(2014)Kholgade, Simon, Efros, and Sheikh]{kholgade20143d}
Natasha Kholgade, Tomas Simon, Alexei Efros, and Yaser Sheikh.
\newblock 3d object manipulation in a single photograph using stock 3d models.
\newblock \emph{ACM Transactions on graphics (TOG)}, 33\penalty0 (4):\penalty0 1--12, 2014.

\bibitem[Kirillov et~al.(2023)Kirillov, Mintun, Ravi, Mao, Rolland, Gustafson, Xiao, Whitehead, Berg, Lo, et~al.]{kirillov2023segment}
Alexander Kirillov, Eric Mintun, Nikhila Ravi, Hanzi Mao, Chloe Rolland, Laura Gustafson, Tete Xiao, Spencer Whitehead, Alexander~C Berg, Wan-Yen Lo, et~al.
\newblock Segment anything.
\newblock In \emph{Proceedings of the IEEE/CVF International Conference on Computer Vision}, pages 4015--4026, 2023.

\bibitem[Ku et~al.(2024)Ku, Wei, Ren, Yang, and Chen]{ku2024anyv2v}
Max Ku, Cong Wei, Weiming Ren, Huan Yang, and Wenhu Chen.
\newblock Anyv2v: A plug-and-play framework for any video-to-video editing tasks.
\newblock \emph{arXiv preprint arXiv:2403.14468}, 2024.

\bibitem[Li et~al.(2023)Li, Tan, Zhang, Xu, Luan, Xu, Hong, Sunkavalli, Shakhnarovich, and Bi]{li2023instant3d}
Jiahao Li, Hao Tan, Kai Zhang, Zexiang Xu, Fujun Luan, Yinghao Xu, Yicong Hong, Kalyan Sunkavalli, Greg Shakhnarovich, and Sai Bi.
\newblock Instant3d: Fast text-to-3d with sparse-view generation and large reconstruction model.
\newblock \emph{arXiv preprint arXiv:2311.06214}, 2023.

\bibitem[Li et~al.(2024)Li, Wang, Zhang, Wang, Yuan, Xie, Zou, and Shan]{li2024image}
Yaowei Li, Xintao Wang, Zhaoyang Zhang, Zhouxia Wang, Ziyang Yuan, Liangbin Xie, Yuexian Zou, and Ying Shan.
\newblock Image conductor: Precision control for interactive video synthesis.
\newblock \emph{arXiv preprint arXiv:2406.15339}, 2024.

\bibitem[Lin et~al.(2024{\natexlab{a}})Lin, Ge, Cheng, Li, Zhu, Wang, He, Ye, Yuan, Chen, et~al.]{lin2024open}
Bin Lin, Yunyang Ge, Xinhua Cheng, Zongjian Li, Bin Zhu, Shaodong Wang, Xianyi He, Yang Ye, Shenghai Yuan, Liuhan Chen, et~al.
\newblock Open-sora plan: Open-source large video generation model.
\newblock \emph{arXiv preprint arXiv:2412.00131}, 2024{\natexlab{a}}.

\bibitem[Lin et~al.(2024{\natexlab{b}})Lin, Cho, Zala, and Bansal]{lin2024ctrl}
Han Lin, Jaemin Cho, Abhay Zala, and Mohit Bansal.
\newblock Ctrl-adapter: An efficient and versatile framework for adapting diverse controls to any diffusion model.
\newblock \emph{arXiv preprint arXiv:2404.09967}, 2024{\natexlab{b}}.

\bibitem[Liu et~al.(2023)Liu, Wu, Van~Hoorick, Tokmakov, Zakharov, and Vondrick]{liu2023zero}
Ruoshi Liu, Rundi Wu, Basile Van~Hoorick, Pavel Tokmakov, Sergey Zakharov, and Carl Vondrick.
\newblock Zero-1-to-3: Zero-shot one image to 3d object.
\newblock In \emph{Proceedings of the IEEE/CVF international conference on computer vision}, pages 9298--9309, 2023.

\bibitem[Ma et~al.(2023)Ma, Lewis, and Kleijn]{ma2023trailblazer}
Wan-Duo~Kurt Ma, John~P Lewis, and W~Bastiaan Kleijn.
\newblock Trailblazer: Trajectory control for diffusion-based video generation.
\newblock \emph{arXiv preprint arXiv:2401.00896}, 2023.

\bibitem[Ma et~al.(2024)Ma, He, Cun, Wang, Chen, Li, and Chen]{ma2024follow}
Yue Ma, Yingqing He, Xiaodong Cun, Xintao Wang, Siran Chen, Xiu Li, and Qifeng Chen.
\newblock Follow your pose: Pose-guided text-to-video generation using pose-free videos.
\newblock In \emph{Proceedings of the AAAI Conference on Artificial Intelligence}, pages 4117--4125, 2024.

\bibitem[Melas-Kyriazi et~al.(2023)Melas-Kyriazi, Laina, Rupprecht, and Vedaldi]{melas2023realfusion}
Luke Melas-Kyriazi, Iro Laina, Christian Rupprecht, and Andrea Vedaldi.
\newblock Realfusion: 360deg reconstruction of any object from a single image.
\newblock In \emph{Proceedings of the IEEE/CVF conference on computer vision and pattern recognition}, pages 8446--8455, 2023.

\bibitem[Michel et~al.(2024)Michel, Bhattad, VanderBilt, Krishna, Kembhavi, and Gupta]{michel2024object}
Oscar Michel, Anand Bhattad, Eli VanderBilt, Ranjay Krishna, Aniruddha Kembhavi, and Tanmay Gupta.
\newblock Object 3dit: Language-guided 3d-aware image editing.
\newblock \emph{Advances in Neural Information Processing Systems}, 36, 2024.

\bibitem[Pandey et~al.(2024)Pandey, Guerrero, Gadelha, Hold-Geoffroy, Singh, and Mitra]{pandey2024diffusion}
Karran Pandey, Paul Guerrero, Matheus Gadelha, Yannick Hold-Geoffroy, Karan Singh, and Niloy~J Mitra.
\newblock Diffusion handles enabling 3d edits for diffusion models by lifting activations to 3d.
\newblock In \emph{Proceedings of the IEEE/CVF Conference on Computer Vision and Pattern Recognition}, pages 7695--7704, 2024.

\bibitem[Podell et~al.(2023)Podell, English, Lacey, Blattmann, Dockhorn, M{\"u}ller, Penna, and Rombach]{podell2023sdxl}
Dustin Podell, Zion English, Kyle Lacey, Andreas Blattmann, Tim Dockhorn, Jonas M{\"u}ller, Joe Penna, and Robin Rombach.
\newblock Sdxl: Improving latent diffusion models for high-resolution image synthesis.
\newblock \emph{arXiv preprint arXiv:2307.01952}, 2023.

\bibitem[Polyak et~al.(2024)Polyak, Zohar, Brown, Tjandra, Sinha, Lee, Vyas, Shi, Ma, Chuang, et~al.]{polyak2024movie}
Adam Polyak, Amit Zohar, Andrew Brown, Andros Tjandra, Animesh Sinha, Ann Lee, Apoorv Vyas, Bowen Shi, Chih-Yao Ma, Ching-Yao Chuang, et~al.
\newblock \emph{arXiv preprint arXiv:2410.13720}, 2024.

\bibitem[Radford et~al.(2021)Radford, Kim, Hallacy, Ramesh, Goh, Agarwal, Sastry, Askell, Mishkin, Clark, et~al.]{radford2021learning}
Alec Radford, Jong~Wook Kim, Chris Hallacy, Aditya Ramesh, Gabriel Goh, Sandhini Agarwal, Girish Sastry, Amanda Askell, Pamela Mishkin, Jack Clark, et~al.
\newblock Learning transferable visual models from natural language supervision.
\newblock In \emph{International conference on machine learning}, pages 8748--8763. PMLR, 2021.

\bibitem[Rombach et~al.(2022)Rombach, Blattmann, Lorenz, Esser, and Ommer]{rombach2022high}
Robin Rombach, Andreas Blattmann, Dominik Lorenz, Patrick Esser, and Bj{\"o}rn Ommer.
\newblock High-resolution image synthesis with latent diffusion models.
\newblock In \emph{Proceedings of the IEEE/CVF conference on computer vision and pattern recognition}, pages 10684--10695, 2022.

\bibitem[Ruiz et~al.(2023)Ruiz, Li, Jampani, Pritch, Rubinstein, and Aberman]{ruiz2023dreambooth}
Nataniel Ruiz, Yuanzhen Li, Varun Jampani, Yael Pritch, Michael Rubinstein, and Kfir Aberman.
\newblock Dreambooth: Fine tuning text-to-image diffusion models for subject-driven generation.
\newblock In \emph{Proceedings of the IEEE/CVF conference on computer vision and pattern recognition}, pages 22500--22510, 2023.

\bibitem[Sargent et~al.(2023)Sargent, Li, Shah, Herrmann, Yu, Zhang, Chan, Lagun, Fei-Fei, Sun, et~al.]{sargent2023zeronvs}
Kyle Sargent, Zizhang Li, Tanmay Shah, Charles Herrmann, Hong-Xing Yu, Yunzhi Zhang, Eric~Ryan Chan, Dmitry Lagun, Li Fei-Fei, Deqing Sun, et~al.
\newblock Zeronvs: Zero-shot 360-degree view synthesis from a single real image.
\newblock \emph{arXiv preprint arXiv:2310.17994}, 2023.

\bibitem[Shi et~al.(2023)Shi, Chen, Zhang, Liu, Xu, Wei, Chen, Zeng, and Su]{shi2023zero123++}
Ruoxi Shi, Hansheng Chen, Zhuoyang Zhang, Minghua Liu, Chao Xu, Xinyue Wei, Linghao Chen, Chong Zeng, and Hao Su.
\newblock Zero123++: a single image to consistent multi-view diffusion base model.
\newblock \emph{arXiv preprint arXiv:2310.15110}, 2023.

\bibitem[Shi et~al.(2024)Shi, Huang, Wang, Bian, Li, Zhang, Zhang, Cheung, See, Qin, et~al.]{shi2024motion}
Xiaoyu Shi, Zhaoyang Huang, Fu-Yun Wang, Weikang Bian, Dasong Li, Yi Zhang, Manyuan Zhang, Ka~Chun Cheung, Simon See, Hongwei Qin, et~al.
\newblock Motion-i2v: Consistent and controllable image-to-video generation with explicit motion modeling.
\newblock In \emph{ACM SIGGRAPH 2024 Conference Papers}, pages 1--11, 2024.

\bibitem[Song et~al.(2020)Song, Meng, and Ermon]{song2020denoising}
Jiaming Song, Chenlin Meng, and Stefano Ermon.
\newblock Denoising diffusion implicit models.
\newblock \emph{arXiv preprint arXiv:2010.02502}, 2020.

\bibitem[Szymanowicz et~al.(2024)Szymanowicz, Insafutdinov, Zheng, Campbell, Henriques, Rupprecht, and Vedaldi]{szymanowicz2024flash3d}
Stanislaw Szymanowicz, Eldar Insafutdinov, Chuanxia Zheng, Dylan Campbell, Jo{\~a}o~F Henriques, Christian Rupprecht, and Andrea Vedaldi.
\newblock Flash3d: Feed-forward generalisable 3d scene reconstruction from a single image.
\newblock \emph{arXiv preprint arXiv:2406.04343}, 2024.

\bibitem[Tang et~al.(2025)Tang, Chen, Chen, Wang, Zeng, and Liu]{tang2025lgm}
Jiaxiang Tang, Zhaoxi Chen, Xiaokang Chen, Tengfei Wang, Gang Zeng, and Ziwei Liu.
\newblock Lgm: Large multi-view gaussian model for high-resolution 3d content creation.
\newblock In \emph{European Conference on Computer Vision}, pages 1--18. Springer, 2025.

\bibitem[Tochilkin et~al.(2024)Tochilkin, Pankratz, Liu, Huang, Letts, Li, Liang, Laforte, Jampani, and Cao]{tochilkin2024triposr}
Dmitry Tochilkin, David Pankratz, Zexiang Liu, Zixuan Huang, Adam Letts, Yangguang Li, Ding Liang, Christian Laforte, Varun Jampani, and Yan-Pei Cao.
\newblock Triposr: Fast 3d object reconstruction from a single image.
\newblock \emph{arXiv preprint arXiv:2403.02151}, 2024.

\bibitem[Tumanyan et~al.(2023)Tumanyan, Geyer, Bagon, and Dekel]{tumanyan2023plug}
Narek Tumanyan, Michal Geyer, Shai Bagon, and Tali Dekel.
\newblock Plug-and-play diffusion features for text-driven image-to-image translation.
\newblock In \emph{Proceedings of the IEEE/CVF Conference on Computer Vision and Pattern Recognition}, pages 1921--1930, 2023.

\bibitem[Wang et~al.(2024{\natexlab{a}})Wang, Zhang, Zou, Zeng, Wei, Yuan, and Li]{wang2024boximator}
Jiawei Wang, Yuchen Zhang, Jiaxin Zou, Yan Zeng, Guoqiang Wei, Liping Yuan, and Hang Li.
\newblock Boximator: Generating rich and controllable motions for video synthesis.
\newblock \emph{arXiv preprint arXiv:2402.01566}, 2024{\natexlab{a}}.

\bibitem[Wang et~al.(2024{\natexlab{b}})Wang, Leroy, Cabon, Chidlovskii, and Revaud]{wang2024dust3r}
Shuzhe Wang, Vincent Leroy, Yohann Cabon, Boris Chidlovskii, and Jerome Revaud.
\newblock Dust3r: Geometric 3d vision made easy.
\newblock In \emph{Proceedings of the IEEE/CVF Conference on Computer Vision and Pattern Recognition}, pages 20697--20709, 2024{\natexlab{b}}.

\bibitem[Wang et~al.(2024{\natexlab{c}})Wang, Yuan, Zhang, Chen, Wang, Zhang, Shen, Zhao, and Zhou]{wang2024videocomposer}
Xiang Wang, Hangjie Yuan, Shiwei Zhang, Dayou Chen, Jiuniu Wang, Yingya Zhang, Yujun Shen, Deli Zhao, and Jingren Zhou.
\newblock Videocomposer: Compositional video synthesis with motion controllability.
\newblock \emph{Advances in Neural Information Processing Systems}, 36, 2024{\natexlab{c}}.

\bibitem[Wang et~al.(2004)Wang, Bovik, Sheikh, and Simoncelli]{wang2004image}
Zhou Wang, Alan~C Bovik, Hamid~R Sheikh, and Eero~P Simoncelli.
\newblock Image quality assessment: from error visibility to structural similarity.
\newblock \emph{IEEE transactions on image processing}, 13\penalty0 (4):\penalty0 600--612, 2004.

\bibitem[Wang et~al.(2023)Wang, Yuan, Wang, Chen, Xia, Luo, and Shan]{wang2023motionctrl}
Zhouxia Wang, Ziyang Yuan, Xintao Wang, Tianshui Chen, Menghan Xia, Ping Luo, and Ying Shan.
\newblock Motionctrl: A unified and flexible motion controller for video generation.
\newblock \emph{arXiv preprint arXiv:2312.03641}, 2023.

\bibitem[Weijie~Kong and Jie~Jiang(2024)]{kong2024hunyuanvideo}
Zijian Zhang Rox Min Zuozhuo Dai Jin Zhou Jiangfeng Xiong Xin Li Bo Wu Jianwei Zhang Kathrina Wu Qin Lin Aladdin Wang Andong Wang Changlin Li Duojun Huang Fang Yang Hao Tan Hongmei Wang Jacob Song Jiawang Bai Jianbing Wu Jinbao Xue Joey Wang Junkun Yuan Kai Wang Mengyang Liu Pengyu Li Shuai Li Weiyan Wang Wenqing Yu Xinchi Deng Yang Li Yanxin Long Yi Chen Yutao Cui Yuanbo Peng Zhentao Yu Zhiyu He Zhiyong Xu Zixiang Zhou Zunnan Xu Yangyu Tao Qinglin Lu Songtao Liu Dax Zhou Hongfa Wang Yong Yang Di Wang Yuhong~Liu Weijie~Kong, Qi~Tian and along with Caesar~Zhong Jie~Jiang.
\newblock Hunyuanvideo: A systematic framework for large video generative models, 2024.

\bibitem[Wu et~al.(2024{\natexlab{a}})Wu, Li, Zeng, Zhang, Zhou, Li, Tong, and Chen]{wu2024motionbooth}
Jianzong Wu, Xiangtai Li, Yanhong Zeng, Jiangning Zhang, Qianyu Zhou, Yining Li, Yunhai Tong, and Kai Chen.
\newblock Motionbooth: Motion-aware customized text-to-video generation.
\newblock \emph{arXiv preprint arXiv:2406.17758}, 2024{\natexlab{a}}.

\bibitem[Wu et~al.(2025)Wu, Li, Gu, Zhao, He, Zhang, Shou, Li, Gao, and Zhang]{wu2025draganything}
Weijia Wu, Zhuang Li, Yuchao Gu, Rui Zhao, Yefei He, David~Junhao Zhang, Mike~Zheng Shou, Yan Li, Tingting Gao, and Di Zhang.
\newblock Draganything: Motion control for anything using entity representation.
\newblock In \emph{European Conference on Computer Vision}, pages 331--348. Springer, 2025.

\bibitem[Wu et~al.(2024{\natexlab{b}})Wu, Rubanova, Kabra, Hudson, Gilitschenski, Aytar, van Steenkiste, Allen, and Kipf]{wu2024neural}
Ziyi Wu, Yulia Rubanova, Rishabh Kabra, Drew~A Hudson, Igor Gilitschenski, Yusuf Aytar, Sjoerd van Steenkiste, Kelsey~R Allen, and Thomas Kipf.
\newblock Neural assets: 3d-aware multi-object scene synthesis with image diffusion models.
\newblock \emph{arXiv preprint arXiv:2406.09292}, 2024{\natexlab{b}}.

\bibitem[Xing et~al.(2025)Xing, Xia, Zhang, Chen, Yu, Liu, Liu, Wang, Shan, and Wong]{xing2025dynamicrafter}
Jinbo Xing, Menghan Xia, Yong Zhang, Haoxin Chen, Wangbo Yu, Hanyuan Liu, Gongye Liu, Xintao Wang, Ying Shan, and Tien-Tsin Wong.
\newblock Dynamicrafter: Animating open-domain images with video diffusion priors.
\newblock In \emph{European Conference on Computer Vision}, pages 399--417. Springer, 2025.

\bibitem[Xu et~al.(2024)Xu, Cheng, Gao, Wang, Gao, and Shan]{xu2024instantmesh}
Jiale Xu, Weihao Cheng, Yiming Gao, Xintao Wang, Shenghua Gao, and Ying Shan.
\newblock Instantmesh: Efficient 3d mesh generation from a single image with sparse-view large reconstruction models.
\newblock \emph{arXiv preprint arXiv:2404.07191}, 2024.

\bibitem[Yang et~al.(2024{\natexlab{a}})Yang, Kang, Huang, Zhao, Xu, Feng, and Zhao]{yang2024depth}
Lihe Yang, Bingyi Kang, Zilong Huang, Zhen Zhao, Xiaogang Xu, Jiashi Feng, and Hengshuang Zhao.
\newblock Depth anything v2.
\newblock \emph{arXiv preprint arXiv:2406.09414}, 2024{\natexlab{a}}.

\bibitem[Yang et~al.(2024{\natexlab{b}})Yang, Hou, Huang, Ma, Wan, Zhang, Chen, and Liao]{yang2024direct}
Shiyuan Yang, Liang Hou, Haibin Huang, Chongyang Ma, Pengfei Wan, Di Zhang, Xiaodong Chen, and Jing Liao.
\newblock Direct-a-video: Customized video generation with user-directed camera movement and object motion.
\newblock \emph{arXiv preprint arXiv:2402.03162}, 2024{\natexlab{b}}.

\bibitem[Yang et~al.(2023)Yang, Zeng, Yuan, and Li]{yang2023effective}
Zhendong Yang, Ailing Zeng, Chun Yuan, and Yu Li.
\newblock Effective whole-body pose estimation with two-stages distillation.
\newblock In \emph{Proceedings of the IEEE/CVF International Conference on Computer Vision}, pages 4210--4220, 2023.

\bibitem[Yenphraphai et~al.(2024)Yenphraphai, Pan, Liu, Panozzo, and Xie]{yenphraphai2024image}
Jiraphon Yenphraphai, Xichen Pan, Sainan Liu, Daniele Panozzo, and Saining Xie.
\newblock Image sculpting: Precise object editing with 3d geometry control.
\newblock In \emph{Proceedings of the IEEE/CVF Conference on Computer Vision and Pattern Recognition}, pages 4241--4251, 2024.

\bibitem[Yin et~al.(2023)Yin, Wu, Liang, Shi, Li, Ming, and Duan]{yin2023dragnuwa}
Shengming Yin, Chenfei Wu, Jian Liang, Jie Shi, Houqiang Li, Gong Ming, and Nan Duan.
\newblock Dragnuwa: Fine-grained control in video generation by integrating text, image, and trajectory.
\newblock \emph{arXiv preprint arXiv:2308.08089}, 2023.

\bibitem[Yu et~al.(2024{\natexlab{a}})Yu, Duan, Herrmann, Freeman, and Wu]{yu2024wonderworld}
Hong-Xing Yu, Haoyi Duan, Charles Herrmann, William~T Freeman, and Jiajun Wu.
\newblock Wonderworld: Interactive 3d scene generation from a single image.
\newblock \emph{arXiv preprint arXiv:2406.09394}, 2024{\natexlab{a}}.

\bibitem[Yu et~al.(2024{\natexlab{b}})Yu, Xing, Yuan, Hu, Li, Huang, Gao, Wong, Shan, and Tian]{yu2024viewcrafter}
Wangbo Yu, Jinbo Xing, Li Yuan, Wenbo Hu, Xiaoyu Li, Zhipeng Huang, Xiangjun Gao, Tien-Tsin Wong, Ying Shan, and Yonghong Tian.
\newblock Viewcrafter: Taming video diffusion models for high-fidelity novel view synthesis.
\newblock \emph{arXiv preprint arXiv:2409.02048}, 2024{\natexlab{b}}.

\bibitem[Zhang et~al.(2024{\natexlab{a}})Zhang, Li, Wan, Wang, and Liao]{zhang2024text2nerf}
Jingbo Zhang, Xiaoyu Li, Ziyu Wan, Can Wang, and Jing Liao.
\newblock Text2nerf: Text-driven 3d scene generation with neural radiance fields.
\newblock \emph{IEEE Transactions on Visualization and Computer Graphics}, 2024{\natexlab{a}}.

\bibitem[Zhang et~al.(2023{\natexlab{a}})Zhang, Rao, and Agrawala]{zhang2023adding}
Lvmin Zhang, Anyi Rao, and Maneesh Agrawala.
\newblock Adding conditional control to text-to-image diffusion models.
\newblock In \emph{Proceedings of the IEEE/CVF International Conference on Computer Vision}, pages 3836--3847, 2023{\natexlab{a}}.

\bibitem[Zhang et~al.(2024{\natexlab{b}})Zhang, Xu, Wang, Lee, Wetzstein, Zhou, and Yang]{zhang20243ditscene}
Qihang Zhang, Yinghao Xu, Chaoyang Wang, Hsin-Ying Lee, Gordon Wetzstein, Bolei Zhou, and Ceyuan Yang.
\newblock 3ditscene: Editing any scene via language-guided disentangled gaussian splatting.
\newblock \emph{arXiv preprint arXiv:2405.18424}, 2024{\natexlab{b}}.

\bibitem[Zhang et~al.(2023{\natexlab{b}})Zhang, Wei, Jiang, Zhang, Zuo, and Tian]{zhang2023controlvideo}
Yabo Zhang, Yuxiang Wei, Dongsheng Jiang, Xiaopeng Zhang, Wangmeng Zuo, and Qi Tian.
\newblock Controlvideo: Training-free controllable text-to-video generation.
\newblock \emph{arXiv preprint arXiv:2305.13077}, 2023{\natexlab{b}}.

\bibitem[Zheng et~al.(2012)Zheng, Chen, Cheng, Zhou, Hu, and Mitra]{zheng2012interactive}
Youyi Zheng, Xiang Chen, Ming-Ming Cheng, Kun Zhou, Shi-Min Hu, and Niloy~J Mitra.
\newblock Interactive images: Cuboid proxies for smart image manipulation.
\newblock \emph{ACM Trans. Graph.}, 31\penalty0 (4):\penalty0 99--1, 2012.

\end{thebibliography}
}

\end{document}